\documentclass[letterpaper, 10pt, conference]{ieeeconf}

\usepackage[utf8]{inputenc}
\usepackage[T1]{fontenc}
\usepackage{acro}
\usepackage{amsfonts}
\usepackage{amsmath}
\usepackage{graphicx}
\usepackage{lmodern}
\usepackage{pgfplots}
\usepackage{pgfplotstable}
\usepackage[separate-uncertainty=true]{siunitx}
\usepackage{subcaption}
\usepackage{tikz}
\usepackage{url}

\IEEEoverridecommandlockouts
\usetikzlibrary{calc, 3d, scopes}
\newcommand*\circled[1]{\tikz[baseline=(char.base)]{\node[shape=circle, draw, inner sep=2pt] (char) {#1};}}

\DeclareAcronym{BN}{short=BN, long=batch normalization}
\DeclareAcronym{MDP}{short=MDP, long=Markov decision process}
\DeclareAcronym{NN}{short=NN, long=neural network}
\DeclareAcronym{RL}{short=RL, long=reinforcement learning}
\DeclareAcronym{PPH}{short=PPH, long=picks per hour}
\DeclareMathOperator*{\argmax}{arg\,max}
\DeclareMathOperator*{\argmin}{arg\,min}

\title{\LARGE \bf
Improving Data Efficiency of \\ Self-supervised Learning for Robotic Grasping
}

\author{Lars Berscheid$^{1}$, Thomas Rühr$^{2}$ and Torsten Kröger$^{1}$%
\thanks{$^{1}$Intelligent Process Automation and Robotics Lab (IPR),
	Karlsruhe Institute of Technology (KIT)
	{\tt\small \{lars.berscheid, torsten\}@kit.edu}}%
\thanks{$^{2}$KUKA Deutschland GmbH,
	{\tt\small thomas.ruehr@kuka.com}}%
}

\begin{document}

\maketitle


\thispagestyle{empty}
\pagestyle{empty}

© 2019 IEEE. Personal use of this material is permitted. Permission from IEEE must be obtained for all other uses, in any current or future media, including reprinting/republishing this material for advertising or promotional purposes, creating new collective works, for resale or redistribution to servers or lists, or reuse of any copyrighted component of this work in other works.

\vspace{5mm}

\begin{abstract}
Given the task of learning robotic grasping solely based on a depth camera input and gripper force feedback, we derive a learning algorithm from an applied point of view to significantly reduce the amount of required training data. Major improvements in time and data efficiency are achieved by: Firstly, we exploit the geometric consistency between the undistorted depth images and the task space. Using a relative small, fully-convolutional neural network, we predict grasp and gripper parameters with great advantages in training as well as inference performance. Secondly, motivated by the small random grasp success rate of around \SI{3}{\%}, the grasp space was explored in a systematic manner. The final system was learned with \num{23000} grasp attempts in around \SI{60}{h}, improving current solutions by an order of magnitude. For typical bin picking scenarios, we measured a grasp success rate of \SI{96.6 \pm 1.0}{\%}. Further experiments showed that the system is able to generalize and transfer knowledge to novel objects and environments.
\end{abstract}

\section{INTRODUCTION}

Grasping is an essential task in today's robotics, but even more so in future applications like large-scale warehouse automation or service robotics. In particular, bin picking is the task of grasping objects out of unsystematic environments like a randomly filled bin. In comparison to general robotic grasping, bin picking emphasizes some difficulties as partially hidden objects and an obstacle-rich environment. The state-of-the-art industrial solution requires the 3D model of the grasped object. It is localized in a segmented point cloud of the bin, its pose is estimated and grasped at taught coordinates \cite{siciliano_springer_2016}. However, this method does neither work with unknown objects, nor does the manual effort of creating 3D models and hand-teaching grasping coordinates scale to many object types.

\begin{figure}[t]
	\centering
	
	\begin{tikzpicture}
	\node[anchor=south west,inner sep=0] (image) at (0,0) {\includegraphics[width=0.37\textwidth]{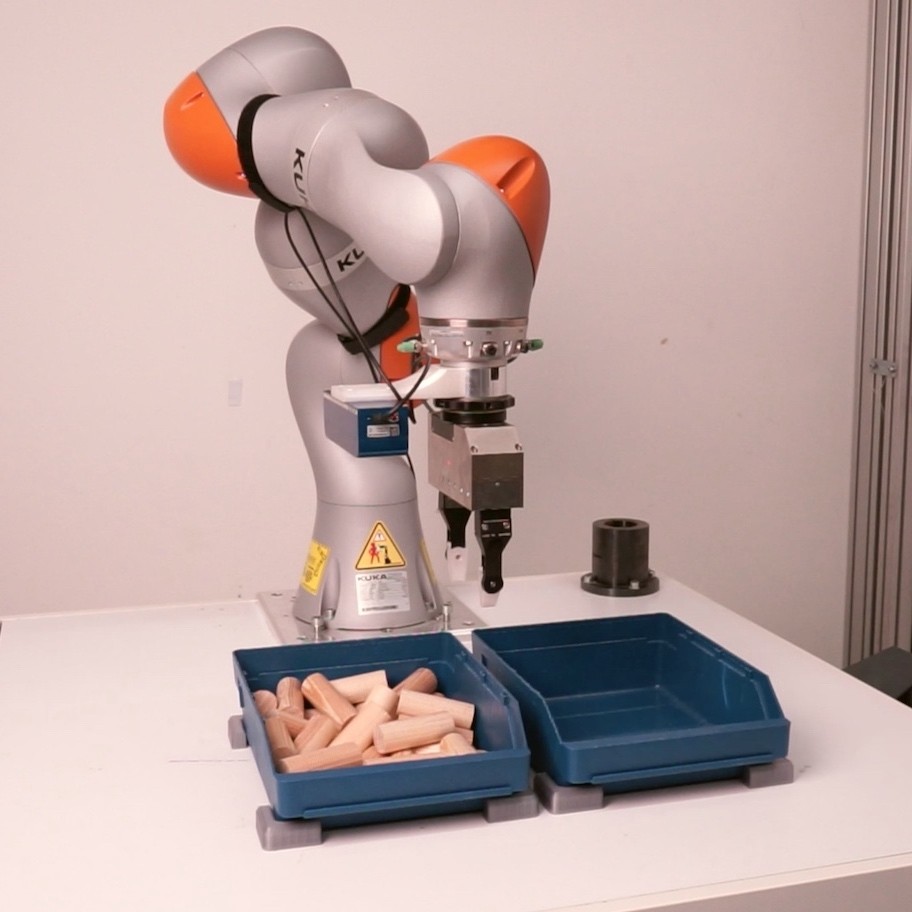}};
	\begin{scope}[x={(image.south east)},y={(image.north west)}]
		\node[] at (0.75, 0.9) {\circled{1}};
		\node[] at (0.15, 0.6) {\circled{2}};
		\node[] at (0.75, 0.55) {\circled{3}};
		\node[] at (0.1, 0.2) {\circled{4}};

		\draw [->] (0.716, 0.87) -- (0.55, 0.75);
		\draw [->] (0.194, 0.59) -- (0.4, 0.54);
		\draw [->] (0.708, 0.535) -- (0.54, 0.48);
		\draw [->] (0.145, 0.204) -- (0.41, 0.23);
	\end{scope}
\end{tikzpicture}

	\caption{Our setup consists of a KUKA LBR iiwa robot (1) with a flange-mounted stereo camera (2), a force-feedback gripper (3), and two industrial storage bins (4).}
	\label{fig:robotic-system}
\end{figure}
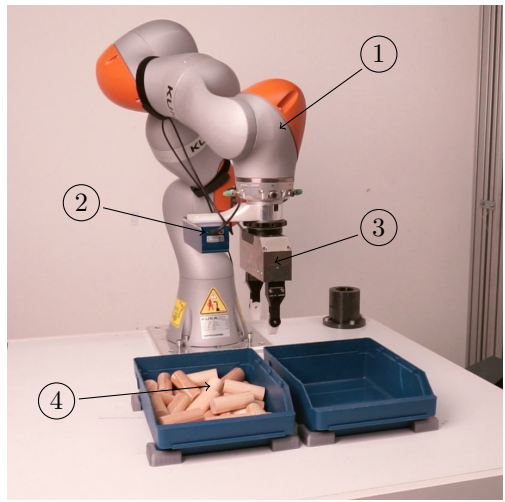

Self-learning methods could reduce this manual input to a minimum. As those data-driven methods have the intrinsic capability to generalize, a universal grasping controller for many objects and scenarios appears possible. Despite recent innovations in robot learning, the progress of self-learning methods is largely based on the consumption of ever-growing datasets. To reverse this costly trend, we emphasize an applied point of view for learning robotic grasping. We therefore consider the time and data efficiency as the major challenge. We restrict ourselves to a simple setup (Fig.~\ref{fig:robotic-system}) with a single robot, typical industrial storage bins and a time frame of a weekend (\SI{60}{h}). Given a depth camera and a force-feedback gripper, the system must be able to learn without human supervision.

Our key contribution is a data-efficient approach for learning robotic grasping. In the following, we will derive our algorithm and simplify where necessary based on the framework of \ac{RL}. The robot trains a \ac{NN} while interacting with the objects using active learning. Therefore, we define a set of rules for exploring the grasp space efficiently. We propose the use of weighted retraining, a procedure for reducing the impact of measurement errors. Then, we present the overall training procedure and implementation. Finally, we conduct real robot experiments, a quantitative grasp rate evaluation and qualitative tests for generalization to unknown objects.

\section{RELATED WORK}

In recent years, research of robotic grasping in unsystematic environments gained a lot of traction. Bohg et al.~\cite{bohg_data-driven_2014} divided possible approaches into the grasping of known, familiar or unknown objects. For known objects, grasping is mostly reduced to object recognition and pose estimation. In a more flexible manner, this process can be extended by matching the similarity of known objects to new, familiar objects. In the case of unknown objects, the approaches can be split into analytical and empirical methods \cite{sahbani_overview_2012}. Analytical grasp synthesis relies on the definition of a grasp quality measure, such as the force closure or other geometrically motivated measures. However, these approaches usually need high-quality and flawless 3D maps of the scene. Data-driven methods intrinsically develop generalization and error robustness, but suffer from the demand of large datasets. First implementations sampled grasp candidates in a simulator and evaluated them based on an analytical measure \cite{miller_graspit!_2004}. In 2008, Saxena et al.~\cite{saxena_robotic_2008} applied a probabilistic model to a mostly simulated dataset to identify grasping points in image patches. More recently, Mahler et al.~\cite{mahler_dex-net_2017} built a large synthetic grasping database with analytic metrics, used supervised learning to train a \ac{NN} and applied it to bin picking \cite{mahler_learning_2017}.

Alternatively to simulation, the system can be automated and trained in a large-scale self-supervised manner. Our work is closely related to the research of Pinto and Gupta~\cite{pinto_supersizing_2016}. They were able to retrain a \ac{NN} within \SI{700}{h} and achieved a grasping rate for seen objects of \SI{73}{\%}. Levine et al.~\cite{levine_learning_2016} scaled up the data collection to \num{800000} grasp attempts on \num{14} robots. They trained in an end-to-end manner, including the camera calibration, spatial relationships and time-dependency.

Our work is based on the theoretical framework of \ac{RL} in combination with convolutional \acp{NN} for function approximation \cite{sutton_reinforcement_1998}. While deep \ac{RL} showed impressive results in learning from visual input in simple simulated environments \cite{rusu_sim_2016, mnih_playing_2013}, direct applications of \ac{RL} for robot learning have proven to be more difficult. On real robots, continuous policy-based methods have been applied to visuomotor learning, either in an end-to-end fashion \cite{levine_end--end_2016} or by using spatial autoencoders \cite{finn_deep_2016}. Quillen et al.~\cite{quillen_deep_2018} showed in a recent comparison of \ac{RL} algorithms for simulated grasping that simple value-based approaches performed best.


\section{LEARNING FOR GRASPING}

\Acf{RL} is a powerful framework to formulate robot learning problems in an abstract way. We define a \ac{MDP} $(\mathcal{S}, \mathcal{A}, T, r, p_0)$ with the state space $\mathcal{S}$, the action space $\mathcal{A}$, the transition distribution $T$, the reward function $r$ and the initial configuration $p_0$. The robot chooses an action $a_t$ at every discrete time step $t$, maximizing the expected collected reward $\mathbb{E} \left[ \sum_t r_t \right]$ over all following time steps. A solution of a \ac{RL} problem is a stochastic policy $\pi: \mathcal{S} \rightarrow \mathcal{P}(\mathcal{A})$, mapping a state $s$ to an action probability distribution $\hat{a}$. A major challenge of \ac{RL} is the possible time-delay between a crucial action $a_t$ and the reward $r_{t + \Delta t}$. To reduce this complexity, we describe a grasp as a single action furthermore. A complete grasp attempt is therefore formulated as sensing a state $s$, planning an action $a$ and receiving a reward $r$ in an open-loop fashion.

\subsection{Single-Shot Grasps}

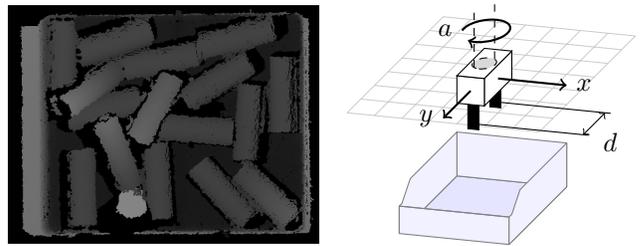
\begin{figure}[t]
	\centering
	
	\begin{subfigure}{0.232\textwidth}
		\vspace{0.2cm}
		\centering
		\includegraphics[trim=105 40 125 40, clip, width=\textwidth]{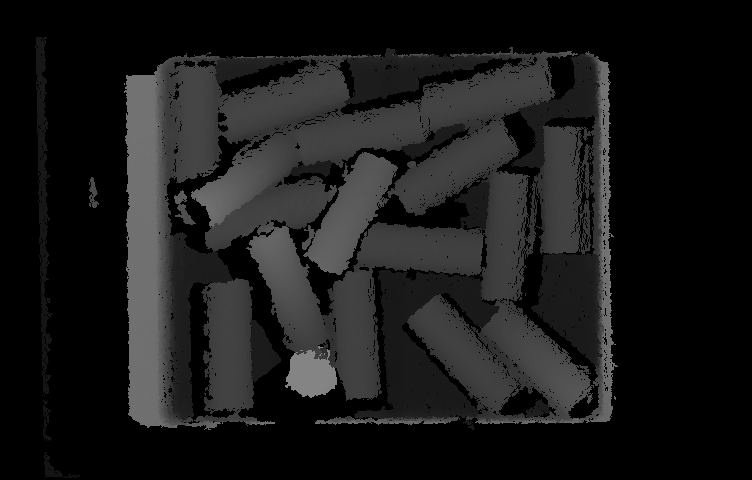}
		\caption{Input depth image}
		\label{fig:example-depth-image}
	\end{subfigure}
	~
	\begin{subfigure}{0.205\textwidth}
		\centering
		\vspace{0.2cm}
		\begin{tikzpicture}[x = {(-0.5cm, -0.5cm)}, y = {(1cm, -0.1cm)}, z = {(0cm, 1cm)}, scale=0.73]
	\definecolor{bincolor}{RGB}{80, 80, 255}
	\tikzstyle{facestyle} = [fill=bincolor!8, draw=gray, thin, line join=round]
	\tikzstyle{ann} = [black, inner sep=1pt]
	\tikzstyle{coordaxis} = [black, thick, ->]
	
	\begin{scope}[canvas is yx plane at z=0.2]
		\path[facestyle, fill=bincolor!15] (-0.5,0) rectangle (1.5,2);
	\end{scope}
	
	\begin{scope}[canvas is zy plane at x=0]
		\path[facestyle] (0.2,-0.5) rectangle (1,1.5);
	\end{scope}
	
	\begin{scope}[canvas is zx plane at y=-0.5]
		\draw[facestyle] (0.2,0) -- (1,0) -- (1,1.7) -- (0.7,2) -- ++(0,0.1) -- (0.2,2.1) -- (0.2,0);
	\end{scope}
	
	\begin{scope}[canvas is zy plane at x=2.1]
		\path[facestyle] (0.2,-0.5) rectangle (0.7,1.5);
	\end{scope}
	
	\begin{scope}[canvas is zx plane at y=1.5]
		\draw[facestyle] (0.2,0) -- (1,0) -- (1,1.7) -- (0.7,2) -- ++(0,0.1) -- (0.2,2.1) -- (0.2,0);
	\end{scope}
	
	\begin{scope}[canvas is yx plane at z=0.7]
		\path[facestyle] (-0.5,2.1) rectangle (1.5,2.1);
	\end{scope}
	
	\begin{scope}[canvas is yx plane at z=2.25]
		\draw[step=5mm, very thin, gray!40] (-1.5,-1.5) grid (2,2);
	\end{scope}

	\node[] (gripper) at (0, 0, 1.8) {};
	\draw[black, fill=black] (gripper)++(0.4,0,0)++(0,0.1,0) -- ++(0,0,-0.5) -- ++(0,-0.2,0) -- ++(0,0,0.5) -- cycle;
	\draw[black, fill=black] (gripper)++(-0.4,0,0)++(0,0.1,0) -- ++(0,0,-0.5) -- ++(0,-0.2,0) -- ++(0,0,0.5) -- cycle;
	\draw[black, fill=white] (gripper)++(-0.5,0.25,0) -- ++(1,0,0) -- ++(0,0,0.5) -- ++(-1,0,0) -- cycle;
	\draw[black, fill=white] (gripper)++(0.5,-0.25,0) -- ++(0,0.5,0) -- ++(0,0,0.5) -- ++(0,-0.5,0) -- cycle;
	\draw[black, fill=white] (gripper)++(-0.5,-0.25,0.5) -- ++(0,0.5,0) -- ++(1,0,0) -- ++(0,-0.5,0) -- cycle;
	\draw[black, thin, fill=gray!30, dashed] (gripper)++(0,0,0.5) circle (0.2);
	\draw[black, thin, dashed] (gripper)++(0.173,-0.1,0.5) -- ++(0,0,1);
	\draw[black, thin, dashed] (gripper)++(-0.173,0.1,0.5) -- ++(0,0,1);
	
	\draw[black, thin] (0.4,0,1.4) -- +(0,2.1,0);
	\draw[black, thin] (-0.4,0,1.4) -- +(0,2.1,0);
	\draw[black, thin, <->] (-0.4,2,1.4) -- ++(0.8,0,0) node[below right, midway] {$d$};
	
	\draw[coordaxis] (0,0.25,2.05) -- +(0,1.25,0) node[right] {$x$};
	\draw[coordaxis] (0.5,0,2.05) -- +(1,0,0) node[left] {$y$};
	\draw[coordaxis, <-] (0.2,-0.2,2.8) arc(-20:280:.4) node[left] {$a$};
		\end{tikzpicture}
		\caption{Output parameters}
		\label{fig:task-space}
	\end{subfigure}
	
	\caption{Our algorithm maps an undistorted depth image (a) to four parameters $(x, y, a, d)$ in the gripper-task space (b). The image ranges from near (white) to far (black), missing depth information are displayed black. The image plane is aligned with the $x$-$y$-plane of the task space, which is a crucial requirement for our algorithm.}
	\label{fig:input-task-space}
\end{figure}

The grasping pose is defined as the position $(x, y, z)$ and Euler angles $(a, b, c)$ (with $a$ around the $z$-axis) where the gripper is closed. Given a calculated grasping pose, the natural process of a single-shot grasp attempt is:
\begin{enumerate}
\item The gripper jaw distance is set to $d$.

\item The gripper approaches the grasping pose with a movement parallel to its fingers. If the joint torque sensors detect a collision, the gripper retracts a few \si{mm} and continues the grasping process.

\item The gripper closes and tries to clamp an object with a predefined force $f$.

\item The robot lifts the object and moves to a filing position. The reward $r$ is set to $1$ for a successful grasp, otherwise to $0$.
\end{enumerate}
We limit ourselves to planar grasps by setting $b=c=0$. The height $z$ is calculated trivially by reading the corresponding value at $(x, y)$ from the depth image. Eventually, the policy needs to calculate four parameters $(x, y, a, d)$ from the depth image (Fig.~\ref{fig:input-task-space}).

\subsection{Geometric Consistency}

A key contribution of this work is to utilize the geometric consistency between the $x$ and $y$ parameter in the task space and the aligned, undistorted depth image. This is made possible by the invariance of the depth image under planar translations of the robotic camera system. However, this demands that the largest parts of the policy computation is translational invariant itself. Fortunately, in \ac{RL}, the policy is often split into two functions $\pi = \sigma \circ \psi$. In our case, $\psi$, called the implicit policy, is a standard value function, such as the state or action value. Then, $\sigma$ selects the final action distribution $\hat{a}$. For greedy policies in the exploitation phase, $\sigma$ is usually the $\max$-function. For exploration, a random term $\varepsilon$ is often added and decreased with experience. A data-efficient exploration is discussed in section~\ref{subsec:grasp-space-exploration}.

In our work, the value function $\psi$ predicts the grasp success probability at the center of a sliding window. In \ac{RL} terms, this corresponds to the state-value of a substate $s^\prime \subset s$. Although the given reward $r$ is binary, predicting the grasping probability is not a classification but a regression task. The grasping success is intrinsically probabilistic, mostly because the \ac{MDP} is in reality only partially observable. Due to the sliding window, the grasp space $\mathcal{A}$ is discretized.

\subsection{Neural Network}

\begin{figure}[t]
	\centering
	\begin{subfigure}{0.5\textwidth}
	\vspace{0.2cm}
\begin{tikzpicture}[scale=0.69, inner sep=0pt, outer sep=2pt, line join=round, font=\scriptsize]
	\newcommand{\layer}[6][]
	{
		\draw[#1] (#2,-#3 / 2,-#4 / 2) -- ++(0,0,#4) -- ++(0,#3,0) -- ++(0,0,-#4) -- cycle;
		\draw[#1] (#2,-#3 / 2,-#4 / 2) -- +(#5,0,0);
		\draw[#1] (#2,#3 / 2,-#4 / 2) -- +(#5,0,0);
		\draw[#1] (#2,-#3 / 2,#4 / 2) -- +(#5,0,0) node [below=0.2cm, midway, align=center]{#6};
		\draw[#1] (#2,#3 / 2,#4 / 2) -- +(#5,0,0);
		\draw[#1] (#2 + #5,-#3 / 2,-#4 / 2) -- ++(0,0,#4) -- ++(0,#3,0) -- ++(0,0,-#4) -- cycle;
	}
	
	\newcommand{\window}[9][]
	{
		\draw[thin] (#2,-#3 / 2 + #6,-#4 / 2 + #7) -- ++(0,0,#4) -- ++(0,#3,0) -- ++(0,0,-#4) -- cycle;
		\draw[thin] (#2,-#3 / 2 + #6,-#4 / 2 + #7) -- ++(#5,0,0) -- ++(0,#3,0) -- ++(-#5, 0, 0) --cycle;
		\draw[thin] (#2,-#3 / 2 + #6,#4 / 2 + #7) -- ++(#5,0,0) -- ++(0,#3,0) -- ++(-#5, 0, 0) --cycle;
		\draw[thin] (#2 + #5,-#3 / 2 + #6,-#4 / 2 + #7) -- ++(0,0,#4) -- ++(0,#3,0) node [left=#1, midway]{#9} -- ++(0,0,-#4) node [above=#1, midway]{#8} -- cycle;
	}
	
	\newcommand{\calc}[7][]
	{
		\draw[#1, thin] (#2,-#3 / 2 + #6,-#4 / 2 + #7) -- ++(0,0,#4) -- ++(0,#3,0) -- ++(0,0,-#4) -- cycle;
		\draw[#1, thin, dashed] (#2,-#3 / 2 + #6,-#4 / 2 + #7) -- #5;
		\draw[#1, thin, dashed] (#2,#3 / 2 + #6,-#4 / 2 + #7) -- #5;
		\draw[#1, thin, dashed] (#2,-#3 / 2 + #6,#4 / 2 + #7) -- #5;
		\draw[#1, thin, dashed] (#2,#3 / 2 + #6,#4 / 2 + #7) -- #5;
	}
	
	\begin{scope}[canvas is yz plane at x=0.07]
		\node[transform shape, rotate=90] (a) {\includegraphics[width=3cm]{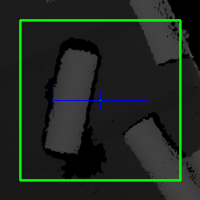}};
	\end{scope}
	
	\layer[thick]{0}{3}{3}{0.07}{1}
	\window[0.1cm, white]{0}{0.6}{0.6}{0.08}{0.5}{-0.3}{5}{5}
	\calc[gray]{0.08}{0.6}{0.6}{(1.4, 0.5, -0.3)}{0.5}{-0.3}
	\node[] () at (-0.2, 1.8) {\textbf{32}};
	\node[] () at (-0.95, -0.4) {\textbf{32}};
	
	\layer[thick]{1.4}{2.8}{2.8}{0.6}{32 \\ Stride $(2, 2)$, \Ac{BN} \\ Dropout $(0.4)$}
	\window[0.1cm]{1.4}{0.6}{0.6}{0.6}{0.5}{-0.3}{5}{5}
	\calc[]{2.0}{0.6}{0.6}{(3.4, -0.2, -0.1)}{0.5}{-0.3}
	
	\layer[thick]{3.4}{1.3}{1.3}{0.9}{48 \\ \Ac{BN} \\ Dropout $(0.5)$}
	\window[0.1cm]{3.4}{0.6}{0.6}{0.9}{-0.2}{-0.1}{5}{5}
	\calc[]{4.3}{0.6}{0.6}{(5.3, -0.1, 0)}{-0.2}{-0.1}
	
	\layer[thick]{5.3}{1.1}{1.1}{1.2}{64 \\ \Ac{BN} \\ \qquad Dropout $(0.6)$}
	\window[0.3cm]{5.3}{0.7}{0.7}{1.2}{-0.1}{0}{6}{6}
	\calc[]{6.5}{0.7}{0.7}{(7.4, 0, 0)}{-0.1}{0}
	
	\layer[thick]{7.4}{0.5}{0.5}{2.7}{142 \\ Dropout $(0.7)$ \\ Regularizer $L_2 = 8.0$}
	\window[0.3cm]{7.4}{0.1}{0.1}{2.7}{0}{0}{1}{1}
	\calc[]{10.2}{0.1}{0.1}{(10.8, 0, 0)}{0}{0}
	
	\layer[thick]{10.8}{0.5}{0.5}{0.1}{3}
	\node[] () at (10.9, 0.6) {\textbf{1}};
	\node[] () at (11.25, 0.0) {\textbf{1}};
\end{tikzpicture}
	\caption{\textbf{Training}}
	\label{fig:nn-training}
	\end{subfigure}
	\begin{subfigure}{0.5\textwidth}
\begin{tikzpicture}[scale=0.69, inner sep=0pt, outer sep=2pt, line join=round, font=\scriptsize]
	\newcommand{\layer}[6][]
	{
		\draw[#1] (#2,-#3 / 2,-#4 / 2) -- ++(0,0,#4) -- ++(0,#3,0) -- ++(0,0,-#4) -- cycle;
		\draw[#1] (#2,#3 / 2,#4 / 2) -- +(#5,0,0);
		\draw[#1] (#2,-#3 / 2,-#4 / 2) -- +(#5,0,0);
		\draw[#1] (#2,#3 / 2,-#4 / 2) -- +(#5,0,0);
		\draw[#1] (#2,-#3 / 2,#4 / 2) -- +(#5,0,0) node [below=0.2cm, midway, align=center]{#6};
		\draw[#1] (#2 + #5,-#3 / 2,-#4 / 2) -- ++(0,0,#4) -- ++(0,#3,0) -- ++(0,0,-#4) -- cycle;
	}
	
	\newcommand{\window}[9][]
	{
		\draw[white, thin] (#2 + #5,-#3 / 2 + #6,-#4 / 2 + #7) -- ++(0,0,#4) -- ++(0,#3,0) node [left=#1, midway]{#9} -- ++(0,0,-#4) node [above=#1, midway]{#8} -- cycle;
	}
	
	\newcommand{\calc}[7][]
	{
		\draw[#1, thin] (#2,-#3 / 2 + #6,-#4 / 2 + #7) -- ++(0,0,#4) -- ++(0,#3,0) -- ++(0,0,-#4) -- cycle;
		\draw[#1, thin, dashed] (#2,-#3 / 2 + #6,-#4 / 2 + #7) -- #5;
		\draw[#1, thin, dashed] (#2,#3 / 2 + #6,-#4 / 2 + #7) -- #5;
		\draw[#1, thin, dashed] (#2,-#3 / 2 + #6,#4 / 2 + #7) -- #5;
		\draw[#1, thin, dashed] (#2,#3 / 2 + #6,#4 / 2 + #7) -- #5;
	}

	\layer[fill=white, rotate around x=-80]{-2.8}{4}{4}{0}{}
	\layer[fill=white, rotate around x=-60]{-2.4}{4}{4}{0}{}
	\layer[fill=white, rotate around x=-40]{-2.0}{4}{4}{0}{}
	\layer[fill=white, rotate around x=-20]{-1.6}{4}{4}{0}{}
	\node[] () at (-1.4, 0.0) {$\cdots$};
	
	\begin{scope}[canvas is yz plane at x=0.07]
		\node[transform shape, rotate=-90] (a) {\includegraphics[width=4cm, height=4cm]{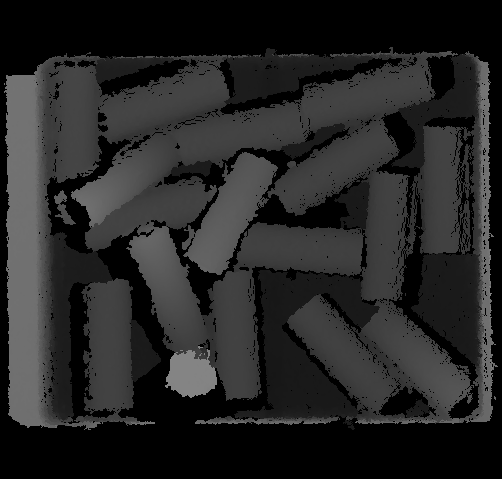}};
	\end{scope}
	
	\draw[decorate, decoration={brace, mirror, raise=10pt}] (-3.6, -2.7) -- (0.8, -2.7) node[midway, below, yshift=-14pt]{20};
	
	\layer[thick]{0}{4}{4}{0.07}{}
	\window[0.1cm, white]{0}{1.2}{1.2}{0.08}{0.6}{-0.3}{32}{32}

	\draw[gray, thin, dashed] (0.08, -1.2 / 2 + 0.6, -1.2 / 2 - 0.3) -- (2.0, -1.2 / 2, -1.2 / 2);
	\draw[gray, thin, dashed] (0.08, -1.2 / 2 + 0.6, 1.2 / 2 - 0.3) -- (2.0, -1.2 / 2, 1.2 / 2);
	\draw[gray, thin, dashed] (0.08, 1.2 / 2 + 0.6, -1.2 / 2 - 0.3) -- (2.0, 1.2 / 2, -1.2 / 2);
	\draw[gray, thin, dashed] (0.08, 1.2 / 2 + 0.6, 1.2 / 2 - 0.3) -- (2.0, 1.2 / 2, 1.2 / 2);
		
	\node[] () at (-0.05, 2.55) {\textbf{110}};
	\node[] () at (1.4, 2.4) {\textbf{110}};
	
	\draw[thick, decorate, decoration={brace,amplitude=6pt}] (2.0, 1) -- (3.6, 1) node[midway, above, yshift=6pt]{\textbf{\Ac{NN} from (a)}};
	
	\layer[thick]{2.0}{1.2}{1.2}{1.2}{}
	\calc[]{3.2}{1.2}{1.2}{(5.0, 0.3, -0.7)}{0}{0}
	
	\layer[thick]{5.0}{2.5}{2.5}{0.2}{3}
	\node[] () at (4.9, 1.6) {\textbf{40}};
	\node[] () at (6.05, 0.5) {\textbf{40}};
	
	\draw[->, thick] (5.6, 1.9) to [out=70, in=130] (7.5, 2.3);
	
	\layer[thick]{7.43}{4}{4}{0.07}{}
	\begin{scope}[canvas is yz plane at x=7.5]
		\node[transform shape, rotate=-90] (a) {\includegraphics[width=4cm, height=4cm]{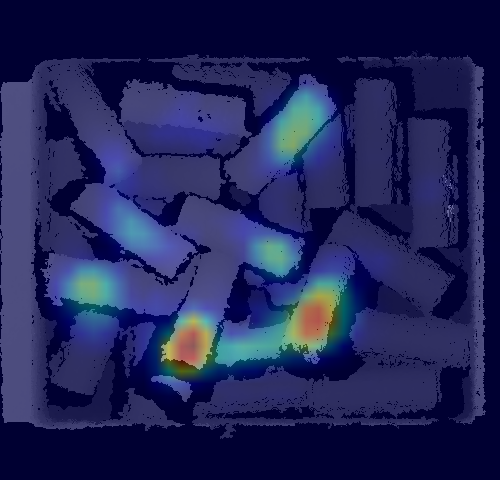}};
	\end{scope}
\end{tikzpicture}
	\caption{\textbf{Inference}}
	\label{fig:nn-inference}
	\end{subfigure}
	
	\caption{(a) Our \acf{NN} architecture. The training output has a size of $(1 \times 1 \times 3)$, corresponding to three gripper distances $d$. (b) During inference, the \ac{NN} calculates from the overview image an output of size $(40 \times 40 \times 3)$, corresponding to the $(x, y, d)$ parameters. For the rotation $a$, the input image is pre-transformed and the \ac{NN} is recalculated for $20$ angles. A single output can be interpreted as a grasp probability heatmap.}
	\label{fig:nn-architecture}
\end{figure}
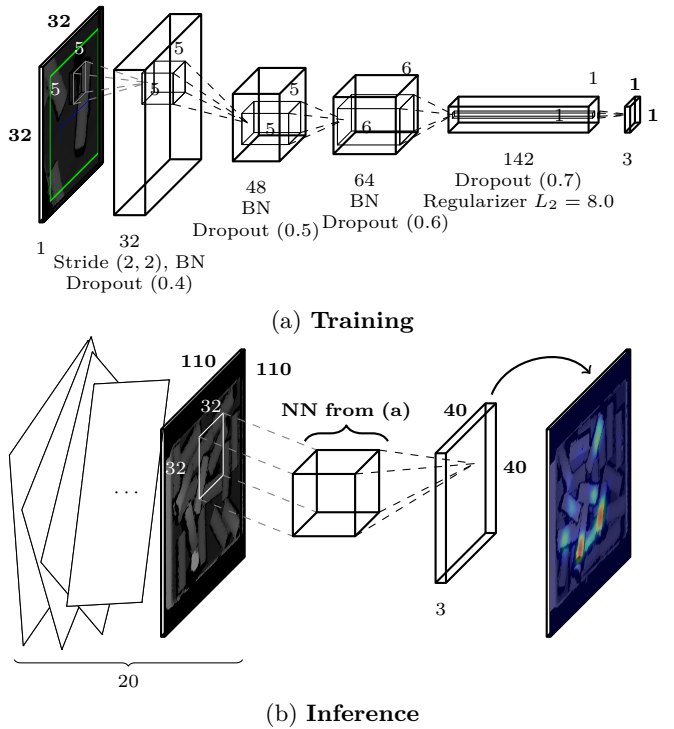

The state-value function $\psi(s^\prime)$, predicting the grasping probability from a depth image subwindow, is approximated by a \ac{NN}. We implement this sliding-window approach as a fully-convolutional \ac{NN}, enabling a variable input and corresponding output size, as proposed by Long et al.~\cite{long_fully_2015}. This increases the inference performance by an order of magnitude. The gripper distance $d$ is integrated as a depth dimension. The \ac{NN} consists of five layers with around \num{450000} parameters (Fig.~\ref{fig:nn-training}). If not otherwise stated, all kernels have single stride and $L_2 = 0.3$. ReLU is used as the activation function. \Acf{BN} and dropout with a minimum rate of $0.4$ are applied to most layers. In comparison to image classification, grasping requires exact spatial information, which would get lost in max-pooling layers. Furthermore, the subwindow images are generated from the overview image using the affine transformation $\mathbf{R}_{a} \mathbf{T}_{xy}$ applying the translation $(x, y)$ and rotation $(a)$ and subsequently cropping around the center. This approach is made possible by the precise intrinsic calibration of the camera as well as the position accuracy of the LBR iiwa, both below a millimeter. Let $\psi(s, a)$ denote the combination of the affine transformation and the state-value function $\psi(s^\prime)$. For training, the loss is defined as the cross entropy of the reward at the measured gripper distance.

During inference, the \ac{NN} scales with the size of the overview depth image (Fig.~\ref{fig:nn-inference}). The angle $a$ is calculated by rotating the input image first, otherwise the kernel sizes would need to be increased accordingly. The grasp space has $\vert \mathcal{A} \vert = \num{96000}$ discrete elements, corresponding to $20$ rotations $a$, $40$ $x$- and $y$ positions and $3$ gripper distances $d$. This leads to a resolution of approximately \SI{4}{mm} in the $x$-$y$-plane, \SI{20}{mm} for the gripper distance $d$ and less than $9^{\circ}$ for the rotation $a$. Calculating a grasp takes around \SI{12}{ms} on a single NVIDIA GeForce GTX 1070, making embedded solutions feasible.

\subsection{Grasp Space Exploration}
\label{subsec:grasp-space-exploration}

\begin{figure*}[t]
	\centering
	\vspace{0.2cm}
	
	\begin{subfigure}{0.13\textwidth}
		\includegraphics[width=\textwidth]{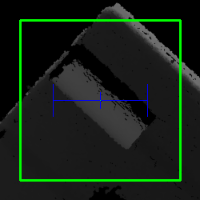}
	\end{subfigure}
	\begin{subfigure}{0.13\textwidth}
		\includegraphics[width=\textwidth]{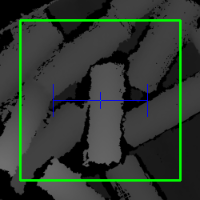}
	\end{subfigure}
	\begin{subfigure}{0.13\textwidth}
		\includegraphics[width=\textwidth]{figures/window-examples/6}
	\end{subfigure}
	\quad
	\begin{subfigure}{0.13\textwidth}
		\includegraphics[width=\textwidth]{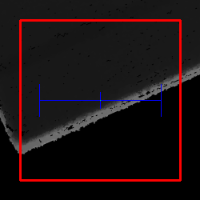}
	\end{subfigure}
	\begin{subfigure}{0.13\textwidth}
		\includegraphics[width=\textwidth]{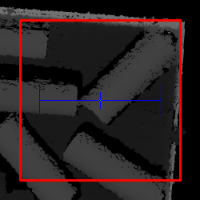}
	\end{subfigure}
	\begin{subfigure}{0.13\textwidth}
		\includegraphics[width=\textwidth]{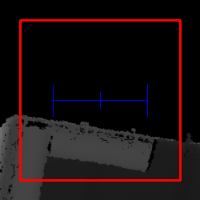}
	\end{subfigure}
	\begin{subfigure}{0.13\textwidth}
		\includegraphics[width=\textwidth]{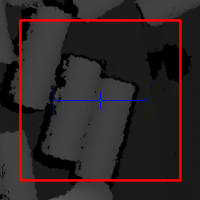}
	\end{subfigure}

	\vspace{0.12cm}

	\begin{subfigure}{0.13\textwidth}
		\includegraphics[width=\textwidth]{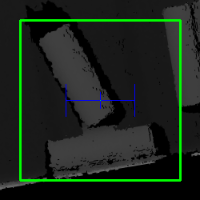}
	\end{subfigure}
	\begin{subfigure}{0.13\textwidth}
		\includegraphics[width=\textwidth]{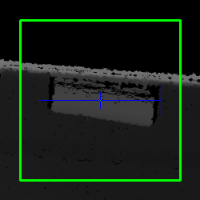}
	\end{subfigure}
	\begin{subfigure}{0.13\textwidth}
		\includegraphics[width=\textwidth]{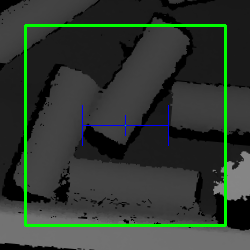}
	\end{subfigure}
	\quad
	\begin{subfigure}{0.13\textwidth}
		\includegraphics[width=\textwidth]{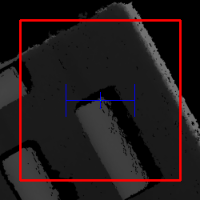}
	\end{subfigure}
	\begin{subfigure}{0.13\textwidth}
		\includegraphics[width=\textwidth]{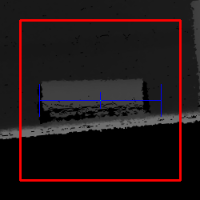}
	\end{subfigure}
	\begin{subfigure}{0.13\textwidth}
		\includegraphics[width=\textwidth]{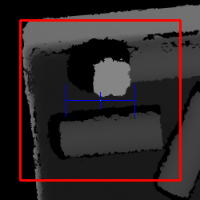}
	\end{subfigure}
	\begin{subfigure}{0.13\textwidth}
		\includegraphics[width=\textwidth]{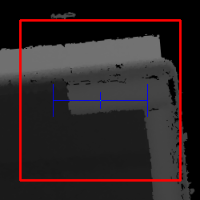}
	\end{subfigure}
	
	\caption{Examples of cropped window depth images. Successful (left, green) and failed (right, red) grasps as well as the gripper position (blue) are marked.}
	\label{fig:window-example-images}
\end{figure*}

Random grasps result in a success rate of only \SI{3}{\%}. This motivates our systematic approach for a data-efficient self-supervised exploration. Therefore, learned information is used to improve learning itself. As this \ac{RL} problem is simplified towards a supervised learning problem, the task of exploration resembles active learning. First of all, images of successful grasps contain more valuable information, as they are harder to collect. Secondary, a grasp attempt is valuable for learning if the predicted grasp probability differs from the measured reward, leading to grasps $a$ given by $\argmax_a \vert r - \psi(s, a) \vert$. To fulfill both conditions, exploration is implemented by choosing one of the following selection functions $\sigma$:

\paragraph{Random}

A uniformly random selection function can be given as $\sigma = \hat{a} \sim P(a) = \vert \mathcal{A} \vert^{-1}$. Only this method is guaranteed to fully span the grasp space $\mathcal{A}$. Random grasp rates were measured using this method.

\paragraph{Maximum(N)}

A more general form of the greedy action $\sigma(\psi) = \argmax_a \psi(s, a)$ is to choose one of the $N$ highest-ranked poses randomly. In application, this method with $N=1$ is first applied. If the grasp fails, a grasp attempt with $N=5$ is repeated. This method is used for exploitation.

\paragraph{Probabilistic}

By choosing a grasp according to the given and normalized probability $\sigma(\psi) = \hat{a} \sim \psi$, the overall measure of probability improves. This method increases the grasping rate but keeps exploration active and is used largely during training.

\paragraph{Uncertain}

The selection function for uncertain grasps with the nearest probability to \num{0.5} is given by $\sigma(\psi) = \argmin_a \vert 0.5 - \psi(s, a) \vert$. Those grasps contain valuable information, because the difference between the grasp probability and the measured reward is near \num{0.5}.

\subsection{Weighted Retraining}

Weighted retraining aims to minimize the impact of two error types. Firstly errors from defective measurements, e.g.\ caused by moving objects, and secondly the sampling error known in \ac{RL}. In the context of a \ac{MDP}, the grasping success is always probabilistic. This is in contrast to the binary value of $r$ and reinforces the sampling error. Let $w_i$ be the weight of a grasp attempt during training and $N$ be the total dataset size. The \ac{NN} is at first trained with $w_i = 1$. Then, the \ac{NN} predicts the reward $\psi$ and adapts the weights according to
\begin{align*}
w_i = \frac{N \left( 1 - \vert r_i - \psi(s_i, a_i) \vert \right)}{\sum_j \left( 1 - \vert r_j - \psi(s_j, a_j) \vert \right)}.
\end{align*}
The normalized difference between measured and predicted reward is used as a quality measure of the input data, and the \ac{NN} is retrained with a weighted training set.

\subsection{Training Procedure}

For training the \ac{NN}, two effects need to be considered: Firstly, an average reward of $0.5$ for the overall dataset is preferred, otherwise the \ac{NN} could overfit to the specific distribution. Secondly, both grasping error types are asymmetric. A false negative error is a missed chance for grasping, while there are typically a multitude of positive grasping poses. However, false positive errors correspond to real grasping errors, which should be minimized with higher priority. Both effects are implemented as two successive, reward-depending weights in the calculation of the cross entropy loss.

To increase the quality of grasps, the clamping force $f$ was reduced during training in comparison to application. This should minimize grasping torques and draw grasps near the center of mass. See \cite{pinto_supervision_2017} for comparison.

The recorded grasp attempts are split into a training and test set. Even as new measurements are added, elements of the two sets must not be exchanged. This is implemented by using a hash function to map the grasp timestamp to a number $\in [0, 1]$. If the value is less than \num{0.2}, the grasp is assigned to the test set. The exact set sizes are therefore not predetermined.

Additionally, our system of two GPUs can train the \ac{NN} as well as infer new grasps for exploration in parallel. Therefore grasp predictions are always up-to-date, improving the data efficiency of the training procedure. The system is able to train completely without human input and supervision.


\section{EXPERIMENTAL RESULTS}

For experiments, a KUKA LBR iiwa \SI{7}{kg} robot arm combined with a Weiss WSG\,50 two-finger gripper was used (Fig.~\ref{fig:robotic-system}). The integrated force-feedback sensor detected if an object was grasped successfully. The flange-mounted Ensenso N10 stereo camera provided 3D data of the scene. If not otherwise stated, wooden cylinders with a diameter of \SI{1.5}{cm} and a height of \SI{6}{cm} were used as test objects. A supplementary video of our experiments can be found at \url{https://youtu.be/0P3FJAo4ZOA}.

\subsection{Data Recording}

\begin{figure*}[h]
	\centering
	\vspace{0.2cm}
	
	\begin{subfigure}{0.19\textwidth}
		\includegraphics[width=\textwidth]{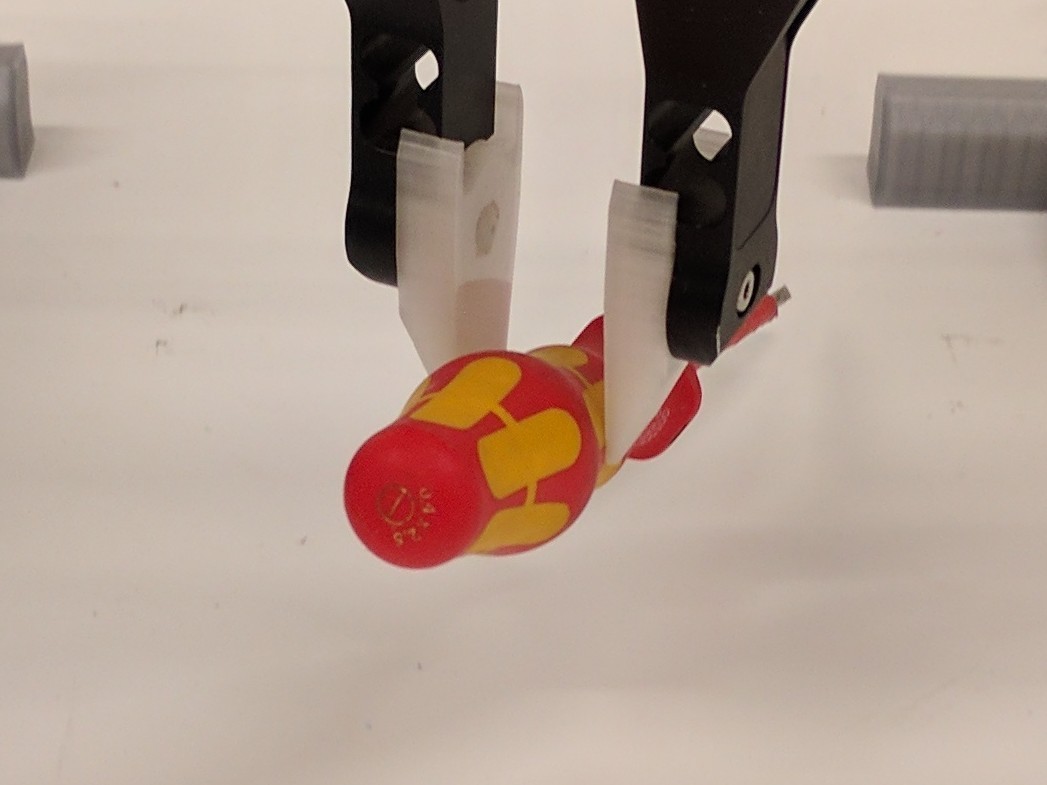}
	\end{subfigure}
	\begin{subfigure}{0.19\textwidth}
		\includegraphics[width=\textwidth]{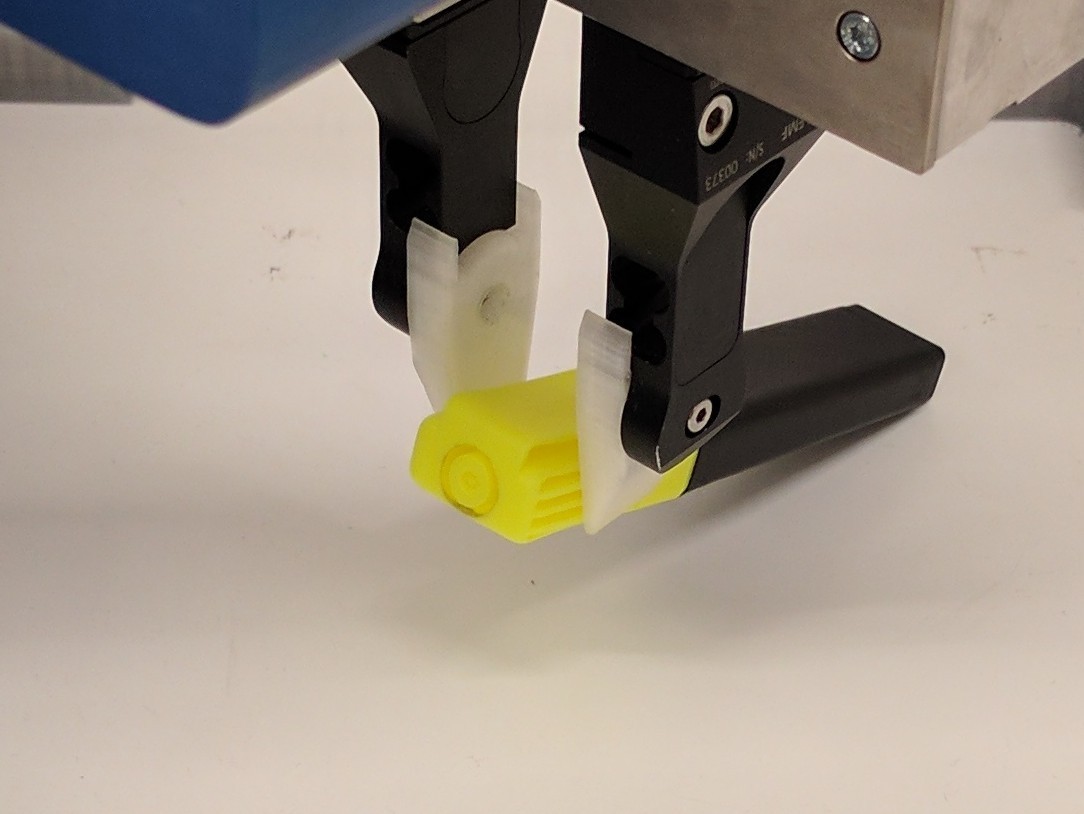}
	\end{subfigure}
	\begin{subfigure}{0.19\textwidth}
		\includegraphics[width=\textwidth]{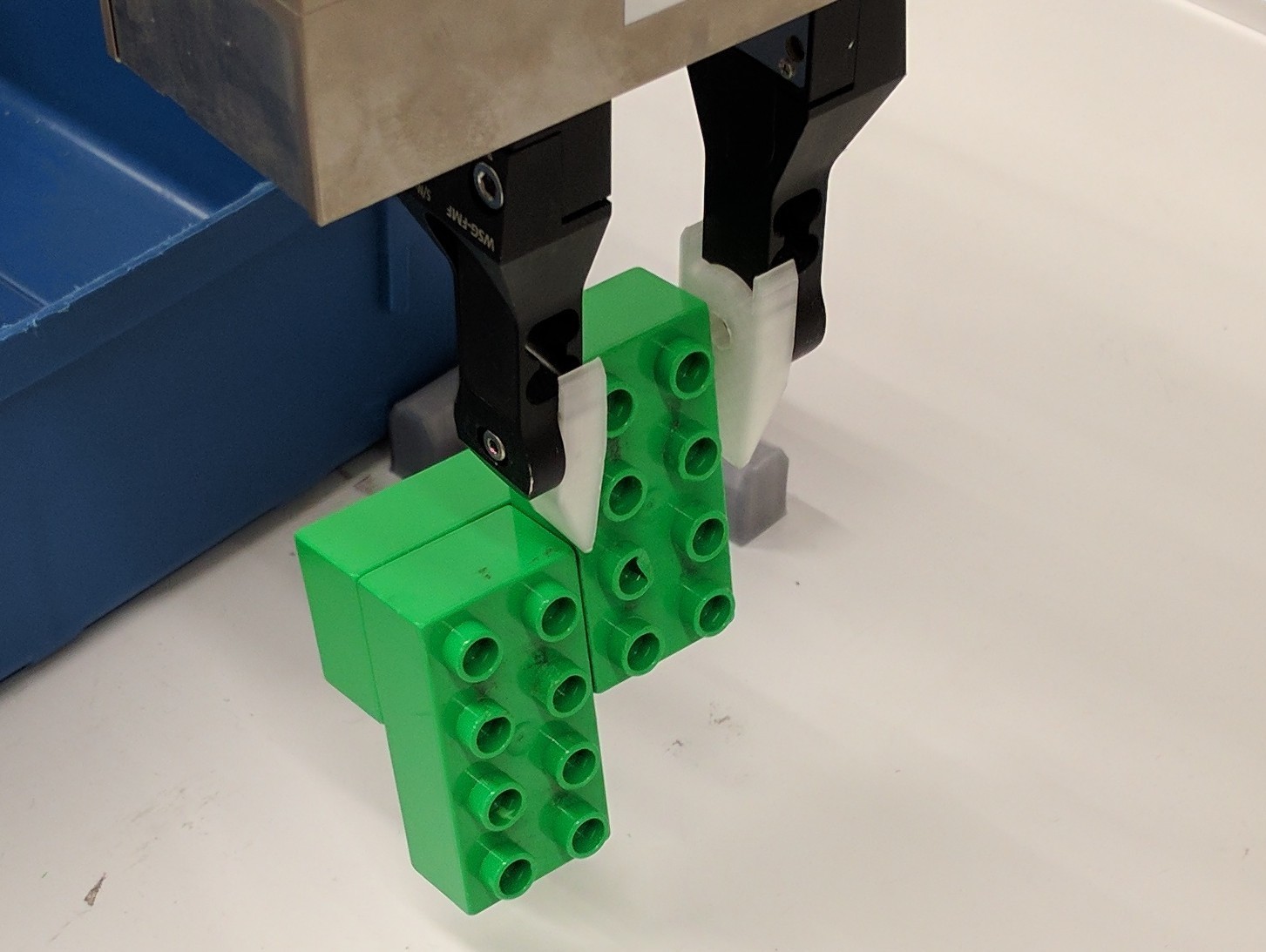}
	\end{subfigure}
	\begin{subfigure}{0.19\textwidth}
		\includegraphics[width=\textwidth]{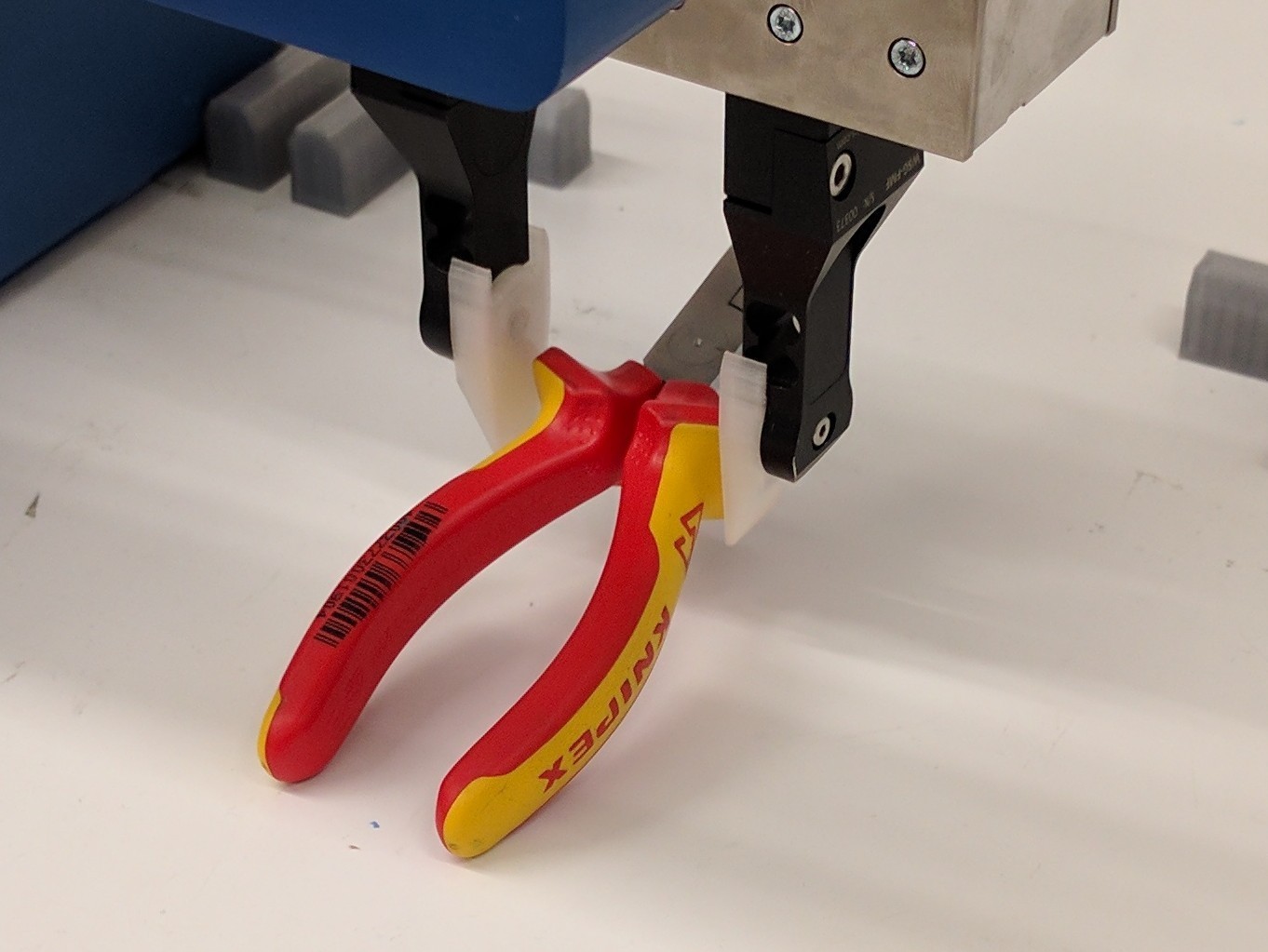}
	\end{subfigure}
	\begin{subfigure}{0.19\textwidth}
		\includegraphics[width=\textwidth]{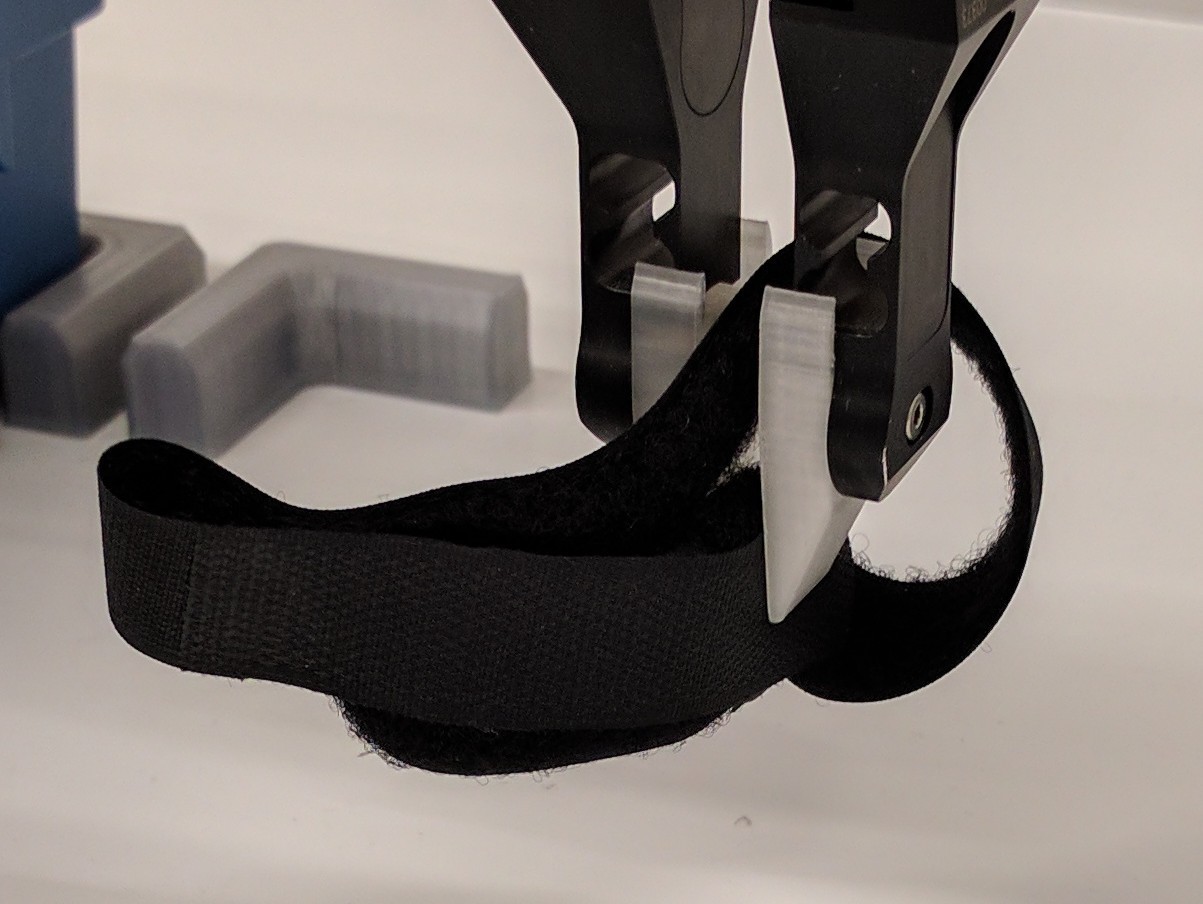}
	\end{subfigure}

	\vspace{0.12cm}

	\begin{subfigure}{0.19\textwidth}
		\includegraphics[width=\textwidth]{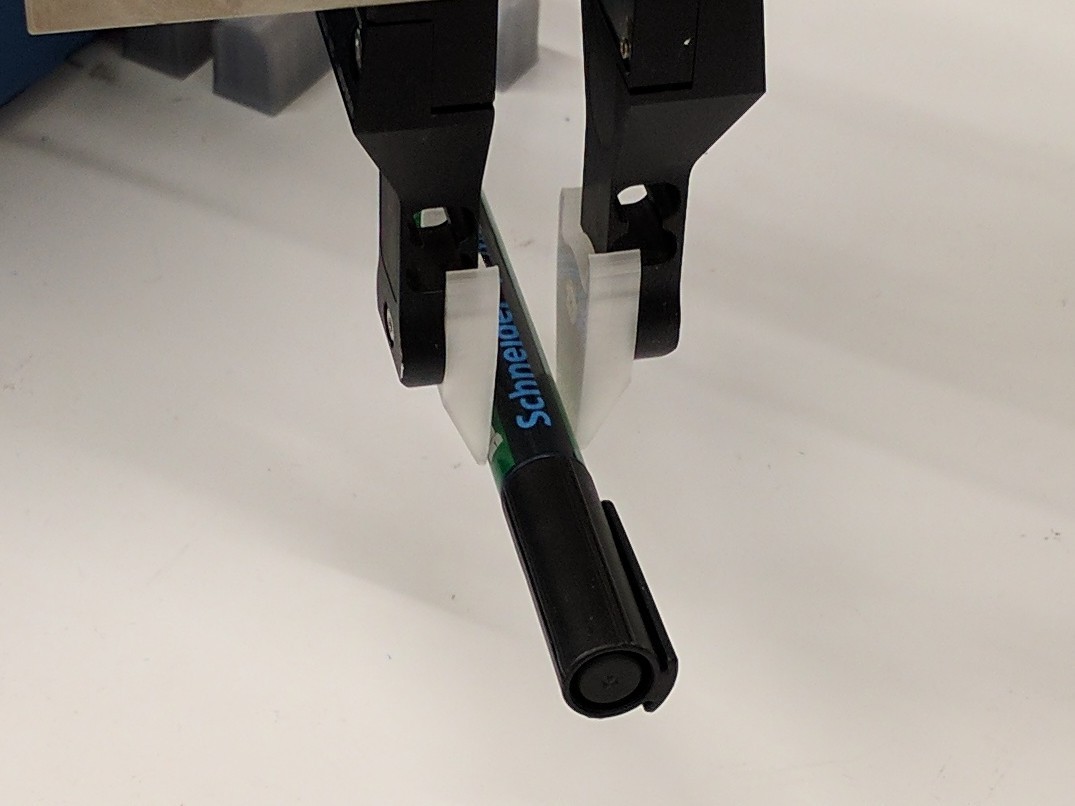}
 	\end{subfigure}
 	\begin{subfigure}{0.19\textwidth}
		\includegraphics[width=\textwidth]{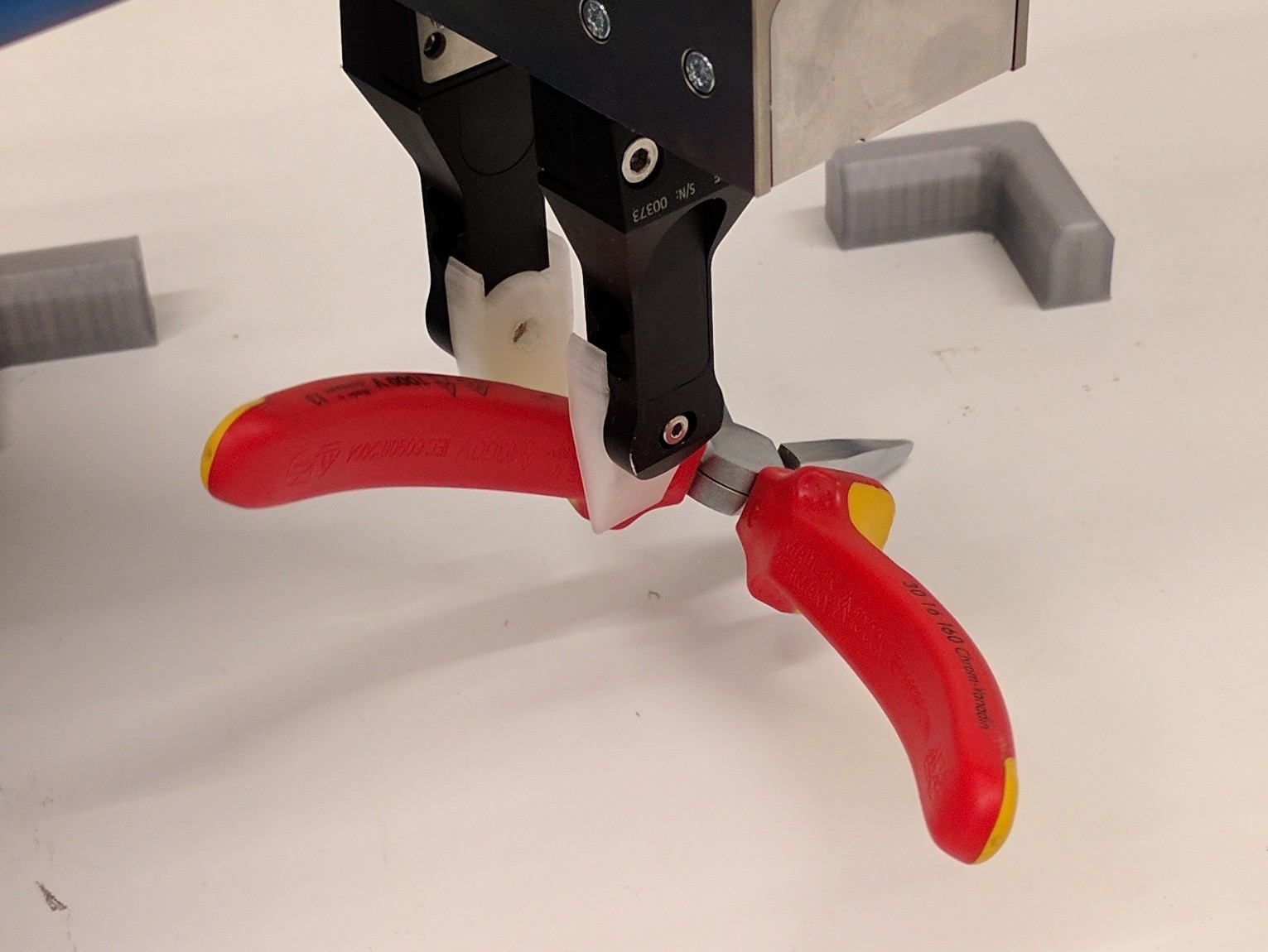}
 	\end{subfigure}
	\begin{subfigure}{0.19\textwidth}
		\includegraphics[width=\textwidth]{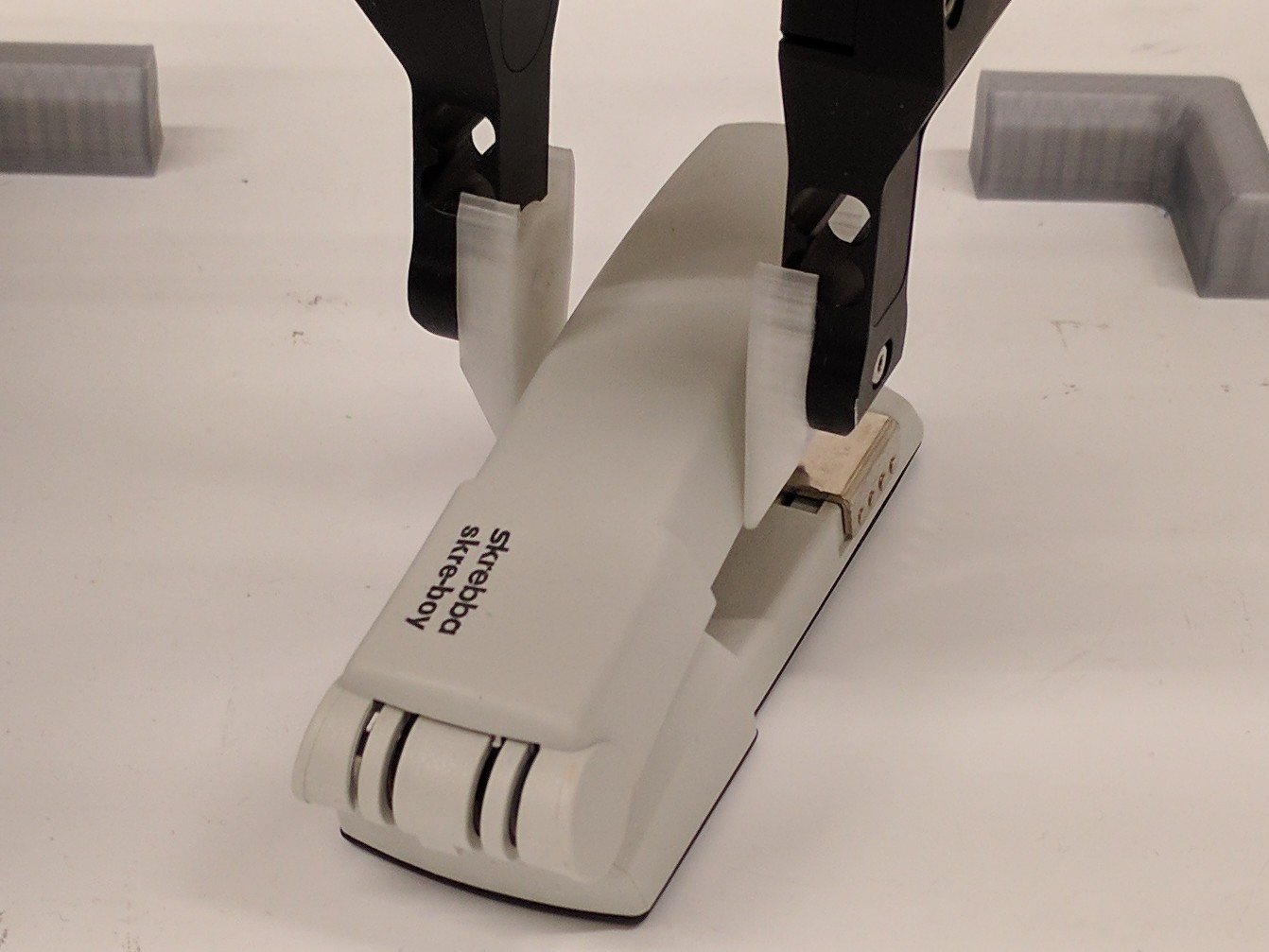}
	\end{subfigure}
	\begin{subfigure}{0.19\textwidth}
		\includegraphics[width=\textwidth]{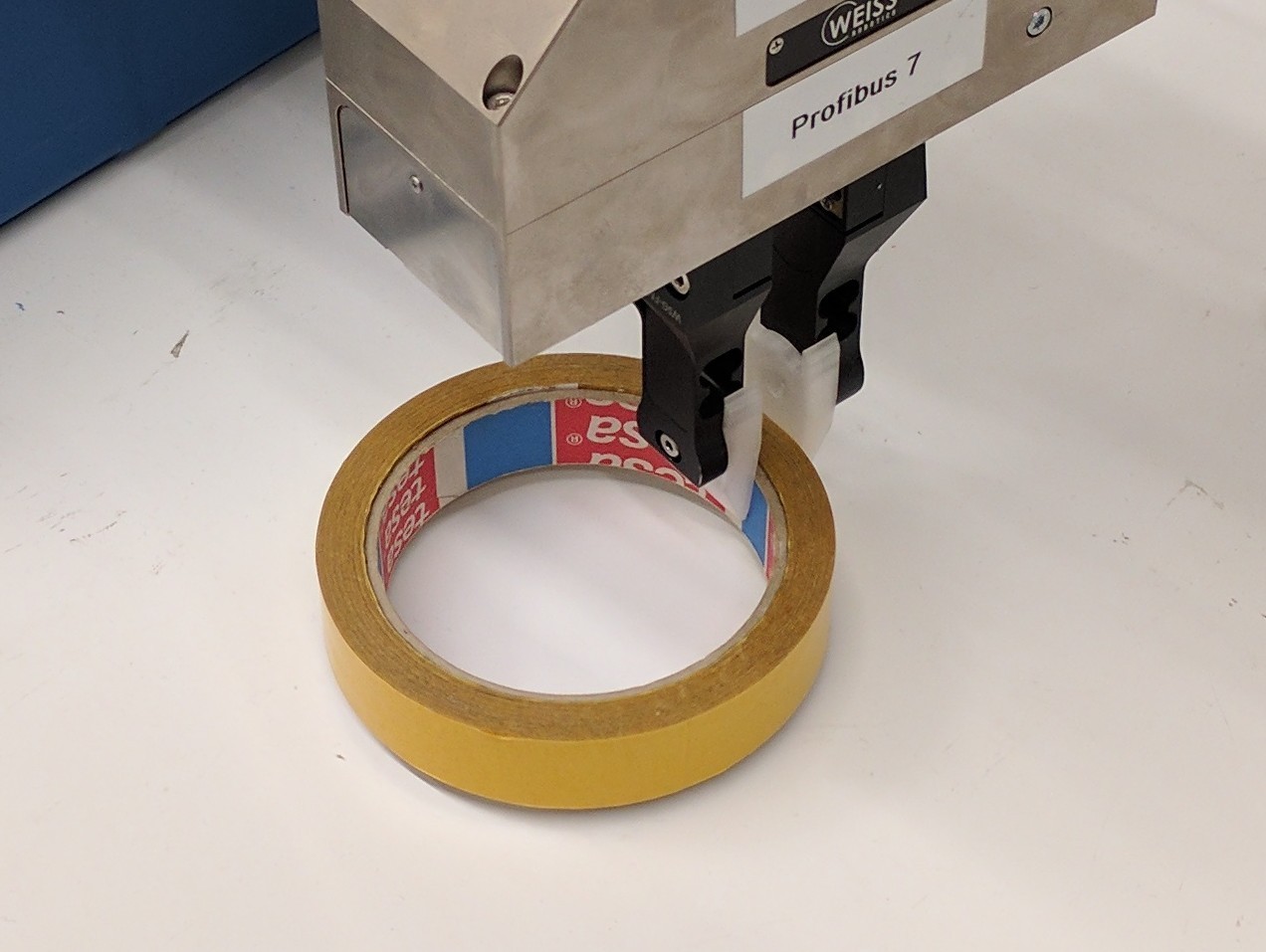}
	\end{subfigure}
 	\begin{subfigure}{0.19\textwidth}
		\includegraphics[width=\textwidth]{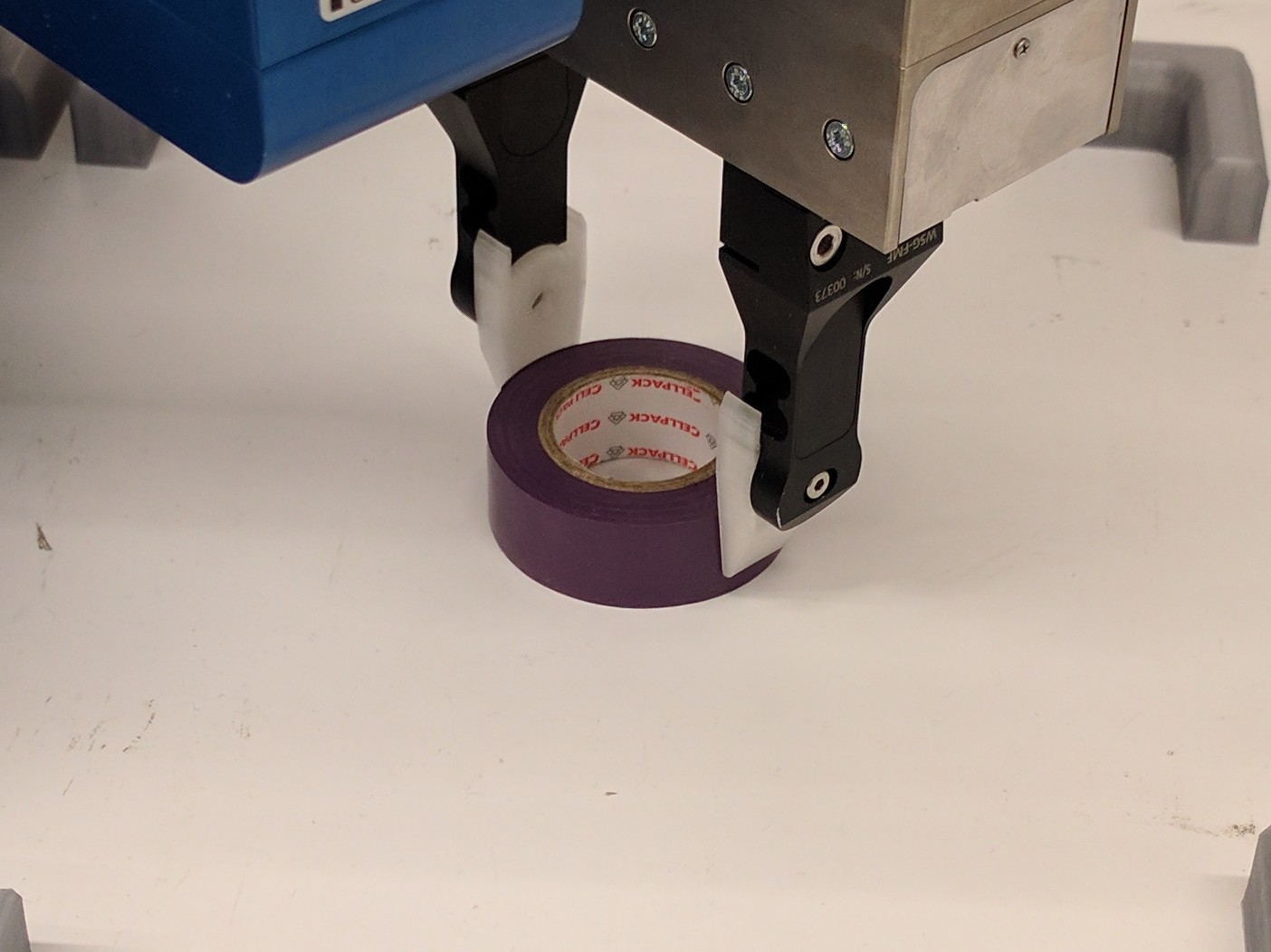}
	\end{subfigure}
	
	\caption{Our robot generalizes to unseen object types. Although it was trained just with cylinders, grasps of similar (left) to more differing (to the right) objects work reliably.}
	\label{fig:novel-objects-images}
\end{figure*}

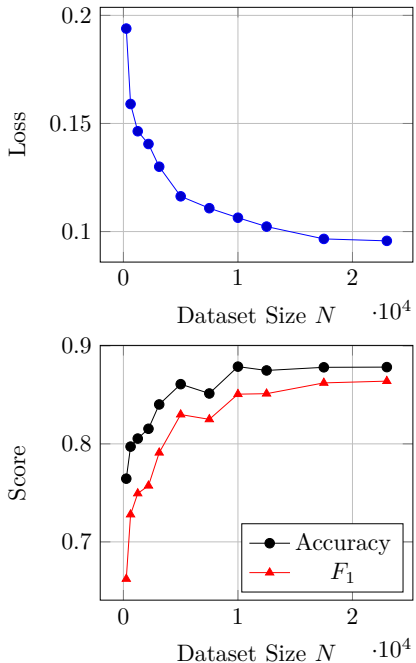
\begin{figure}[t]
	\centering
	
	\begin{subfigure}{0.48\textwidth}
		\centering
\begin{tikzpicture}[scale=0.9]
	\begin{axis}[xlabel=Dataset Size $N$, ylabel=Loss, legend pos=north east, grid=major, width=6.2cm]
		\addplot table [y=Loss, x=N]{data/trainingset-size.txt};
	\end{axis}
\end{tikzpicture}
	\end{subfigure}
	\begin{subfigure}{0.48\textwidth}
		\centering
\begin{tikzpicture}[scale=0.9]
	\begin{axis}[xlabel=Dataset Size $N$, ylabel=Score, legend pos=south east, grid=major, width=6.2cm]
		\addplot [mark=*] table [y=Accuracy, x=N]{data/trainingset-size.txt};
		\addlegendentry{Accuracy}
		
		\addplot [color=red, mark=triangle*] table [y=Fscore, x=N]{data/trainingset-size.txt};
		\addlegendentry{$F_1$}
	\end{axis}
\end{tikzpicture}
	\end{subfigure}

	\caption{Loss, accuracy and $F_1$-score of the test set depending on the dataset size $N$.}
	\label{fig:error-vs-training-set}
\end{figure}

The grasped objects are filed into a second bin. This allows grasp attempts from bins with a varying number of objects and continuous operation by switching grasping and filing bin. Typical durations for grasp attempts are around \SI{10}{s}. The further evaluated \ac{NN} was trained with a dataset of \num{23000} grasp attempts, recorded in \SI{60}{h}. Fig.~\ref{fig:window-example-images} shows examples of the depth image subwindow. Complete random grasps were conducted for the first \num{1000} grasp attempts. During exploration, the random method was used \num{2440}, the probabilistic method \num{8830}, the uncertain method \num{1300} and the maximum(N=5) method \num{9430}~times. Dependencies of different evaluation metrics of the \ac{NN} on the dataset size are shown in Fig.~\ref{fig:error-vs-training-set}.
Note that this development is influenced heavily by the current exploration strategy, since the underlying data distribution changes on purpose. The training was finished as the loss saturated above \num{20000} grasp attempts.

\subsection{Grasp Rate}

The grasp rate is measured by counting the attempts of grasping $n$ objects out of a bin with a total of $m$ objects without replacement. Note that this is more influenced by outliers as well as more important for applications than just counting grasped objects at $n$ attempts. For typical bin picking scenarios, a grasp rate of \SI{96.6 \pm 1.0}{\%} was measured (Table~\ref{tab:results}). Human performance is around \num{600}~\acf{PPH} \cite{mahler_learning_2017}, exceeding our robot system by \SI{40}{\%}. As expected, the grasp rate improves with decreasing clutter and increasing grasp choices.

\begin{table}[h]
	\centering%
	\caption{The measured grasp rate for grasping $n$ objects without replacement out of a bin initially containing $m$ objects. The random grasp rate is $\approx$ \SI{3}{\%}.}
	\begin{tabular}{|c|c|c|c|}
	\hline
	$n$ out of $m$ & Grasp Rate & \ac{PPH} & Number Samples \\ 
	\hline
	1 out of 1 & \SI{99.8 \pm 0.2}{\%} & \num{367 \pm 2} & 460 \\ 
	5 out of 5 & \SI{99.3 \pm 0.7}{\%} & \num{418 \pm 8} & 146 \\ 
	5 out of 10 & \SI{99.5 \pm 0.5}{\%} & \num{428 \pm 3} & 181 \\ 
	10 out of 20 & \SI{96.6 \pm 1.0}{\%} & \num{425 \pm 9} & 207 \\
	\hline
	\end{tabular}
	\label{tab:results}
\end{table}

Weighted retraining of the \ac{NN} improves the grasp rate (\num{10} out of \num{20}) by around \SI{1}{\%} from \SI{95.2 \pm 2.8}{\%} to \SI{96.6 \pm 1.0}{\%}. The main effect of weighted retraining is the reinforcement of reliable grasps. As a side effect, this decreases the probability of making the same grasp mistake again, leading to a reduced uncertainty.

\subsection{Generalization}

While the robot was only trained with cylinder objects, it is able to generalize to novel object types (Fig.~\ref{fig:novel-objects-images}). The grasp rate for the shown isolated objects is above \SI{90}{\%}. Even objects with holes or a flexible geometry work reliably. For cluttered situations, the grasp rate decreases due to stacked and partially hidden objects. Problems arise with objects larger than the trained cylinders. As larger objects may look like multiple smaller objects, the system attempts a grasp to separate possible objects.

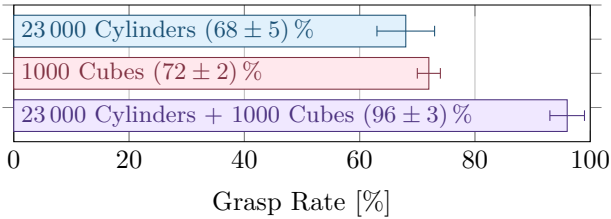
\begin{figure}[h]
	\centering
	\begin{tikzpicture}
\definecolor{clr1}{RGB}{153, 102, 255}
\definecolor{clr2}{RGB}{255, 99, 132}
\definecolor{clr3}{RGB}{54, 162, 235}

\begin{axis}[
	xbar=4pt,
	xmajorgrids,
	bar width=1.2em, 
	width=0.52\textwidth,
	height=3.4cm,
	xmin = 0,
	xmax = 100,
	yticklabels={,,},
	xlabel={Grasp Rate [\%]}, 
	]
\addplot[color=black!50!clr1, fill=clr1!15, error bars/.cd, x dir=both,x explicit] coordinates {(96,0) +- (3, 0)};
\addplot[color=black!50!clr2, fill=clr2!15, error bars/.cd, x dir=both,x explicit] coordinates {(72,0) +- (2, 0)};
\addplot[color=black!50!clr3, fill=clr3!15, error bars/.cd, x dir=both,x explicit] coordinates {(68,0) +- (5, 0)};

\coordinate (A) at (4.1cm, 1.6em);
\coordinate (B) at (4.1cm, 0.0em);
\coordinate (C) at (4.1cm, -1.62em);
\end{axis}
\node[align=left, text width=8cm, font=\small, color=black!50!clr3] at (A) {\num{23000} Cylinders \SI{68 \pm 5}{\%}};
\node[align=left, text width=8cm, font=\small, color=black!50!clr2] at (B) {\num{1000} Cubes \SI{72 \pm 2}{\%}};
\node[align=left, text width=8cm, font=\small, color=black!50!clr1] at (C) {\num{23000} Cylinders + \num{1000} Cubes \SI{96 \pm 3}{\%}};
\end{tikzpicture}
	\caption{The grasp rate of cubes (\num{10} out of \num{15} objects), depending on the trained object type (label) and their corresponding dataset sizes $N$.}
	\label{fig:grasp-rate-cubes}
\end{figure}

The grasping success depends on the similarity with the trained object type. A typical grasp mistake is a collision with sharp edges of objects, which is probably due to the rounded edges of the trained cylinders. Furthermore, the ability to generalize was evaluated with sharp cubes as the novel object type. Fig.~\ref{fig:grasp-rate-cubes} shows the robot's capability of transfer learning between object types. Only by combining the small dataset of cubes with the previous dataset of cylinders, a comparable grasp rate to Table~\ref{tab:results} is achieved. Therefore, the training of novel object types in applications should rely on existing datasets.

The robotic system is also able to generalize to new environments. Simple height differences have no measurable effect, as well as removing the bin and grasping from table surfaces. Tilted backgrounds have only small effects, as the sliding window only operates on local areas of the overall depth image. Even horizontal shelf picking works reliably as well, although the system was trained while gravity pointed in another direction.

\section{DISCUSSION AND OUTLOOK}

We developed a self-learning robotic grasping system while keeping more realistic requirements for industrial applications in mind. In this regard, we used a depth camera in contrast to~\cite{pinto_supersizing_2016, levine_learning_2016, jang_end--end_2017}. Our implementation works on a simple experimental setup and improves time and data efficiency significantly \cite{pinto_supersizing_2016, levine_learning_2016}. An average grasp lasts less than \SI{10}{s}, leading to a fivefold reduction compared to \cite{pinto_supersizing_2016} with a similar setting. Our algorithm runs at \num{50} frames per second and outperforms \cite{redmon_real-time_2015}, despite their focus on real-time capability. Although our random grasp rate of around \SI{3}{\%} is an order of magnitude lower, the final dataset is only half in size in comparison to \cite{pinto_supersizing_2016}. In contrast, we used only one object type in training and testing. For seen objects, \cite{pinto_supersizing_2016} measured a grasp rate of \SI{73}{\%}. \cite{levine_learning_2016} achieved a grasp rate of \SI{90}{\%} for unseen objects in a otherwise similar setting. However, both setups had significantly higher random grasp rates of up to~\SI{33}{\%}. Because we used typical storage bins used in intra-logistic applications, our gripper collided more often during exploration. Consistently, our \ac{NN} additionally needed to learn a collision model between the gripper and the bin. In opposition to~\cite{levine_learning_2016}, our approach generalizes to unseen environments and can therefore be applied not only to bin picking, but to general planar robotic grasping. \cite{pinto_supervision_2017} suggested shaking the grasped object to improve the stability of learned grasps. As an alternative, we simply reduced the gripper force $f$ during training. This leads to similar results, but faster training and less robotic wear and tear.

Due to the planar simplification, the robot can physically not grasp in the corners of the bin. In future, this can be resolved by adding the two remaining angles $b$ and $c$. This increases the dimension of the grasp space, so that a trivial extension of this work would suffer from the curse of dimensionality. We suggest to combine our value-based approach for $x$, $y$ and $a$ with a continuous policy-based method for $b$ and $c$. As more parameters are introduced, the random grasp rate decreases even further. Two main approaches could solve this problem: Firstly, the grasp space exploration should be investigated in greater detail. Secondly, the difficulty of the robotic grasping tasks can be increased in incremental steps. At a more general level, we believe that the robot is able to transfer knowledge from an easier to a more difficult task. In practice, full grasps could be trained by planar grasps while slowly deviating the angles $b$ and $c$. Then, a key challenge in robot learning lies in an appropriate task design.

\section*{ACKNOWLEDGMENT}

This work was performed at KUKA Corporate Research. We would like to thank Pascal Meißner for his helpful feedback and suggestions.

\bibliographystyle{IEEEtran}
\bibliography{root}

\end{document}